\documentclass[10pt,twocolumn,letterpaper]{article}

\usepackage{iccv}
\usepackage{times}
\usepackage{epsfig}
\usepackage{graphicx}
\usepackage{amsmath}
\usepackage{amssymb}
\usepackage{enumitem}
\usepackage{booktabs}
\usepackage{multirow}
\usepackage{etoolbox}
\usepackage{mathtools, cuted}
\usepackage{lipsum}
\usepackage[font=small]{caption}
\usepackage{caption, subcaption, multirow, overpic, textpos}
\usepackage[british, english, american]{babel}

\usepackage[pagebackref=true,breaklinks=true,letterpaper=true,colorlinks,bookmarks=false]{hyperref}
\newcommand{\mypar}[1]{\vspace{-3mm}\paragraph{#1}}
\newcommand{\mysection}[1]{\vspace{-2mm}\section{#1}\vspace{-1mm}}
\newcommand{\mysubsection}[1]{\vspace{-1mm}\subsection{#1}\vspace{-.75mm}}
\newcommand{\mysubsubsection}[1]{\vspace{-2mm}\subsubsection{#1}\vspace{-.25mm}}

\iccvfinalcopy % *** Uncomment this line for the final submission

\def\iccvPaperID{1309} % *** Enter the ICCV Paper ID here

\def\projecturl{https://fredfyyang.github.io/vision-from-touch/}

\hypersetup{
    colorlinks=true,
    citecolor=blue,
}

\begin{document}

%%%%%%%%% TITLE
\title{Generating Visual Scenes from Touch}
\vspace{-1mm}
\author{Fengyu Yang
\qquad
Jiacheng Zhang
\qquad
Andrew Owens  \vspace{3mm} \\
University of Michigan \\
}

\maketitle

\begin{strip}
\centering
    \centering
    \raggedright
    \vspace{-8mm} %
    \includegraphics[width=\textwidth]{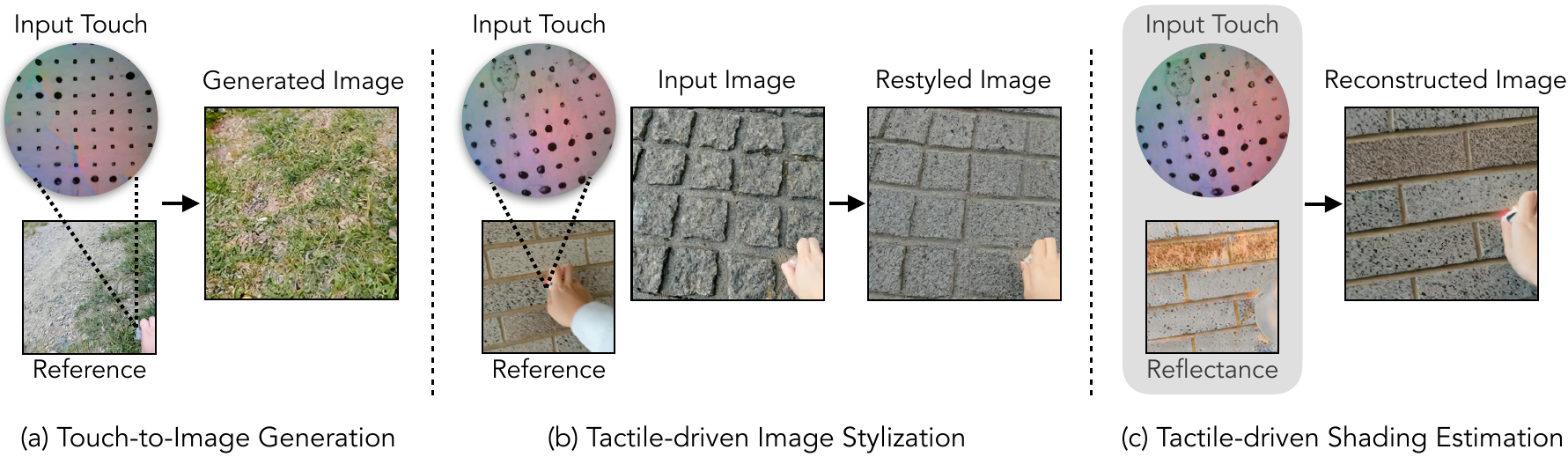}
    \vspace{-6mm}
    \captionof{figure}{\textbf{Generating and manipulating images via touch}. We propose a model based on latent diffusion that translates between touch and images (and vice versa), unifying many previous visuo-tactile image synthesis tasks and enabling new ones. (a) We generate an image of a scene given a tactile signal. (b) We perform tactile-driven image stylization, \eg restyling a rough rock to match the smoother texture of a brick. (c) We propose the novel task of {\em tactile-driven shading estimation}: predicting an image from its reflectance and tactile signal. To aid visualization, we show reference images next to the touch signal. We present a circular crop from the touch signal to emphasize the part of the signal that is in contact with the object.}
  
  \vspace{-3mm}
\label{fig:teaser}
\end{strip}

%%%%%%%%% ABSTRACT
\begin{abstract}
\vspace{-3mm}
An emerging line of work has sought to generate plausible imagery from touch. Existing approaches, however, tackle only narrow aspects of the visuo-tactile synthesis problem, and lag significantly behind the quality of cross-modal synthesis methods in other domains. We draw on recent advances in latent diffusion to create a model for synthesizing images from tactile signals (and vice versa) and apply it to a number of visuo-tactile synthesis tasks. Using this model, we significantly outperform prior work on the tactile-driven stylization problem, i.e., manipulating an image to match a touch signal, and we are the first to successfully generate images from touch without additional sources of information about the scene. We also successfully use our model to address two novel synthesis problems: generating images that do not contain the touch sensor or the hand holding it, and estimating an image's shading from its reflectance and touch. Project Page: \url{\projecturl}

\end{abstract}

%%%%%%%%% BODY TEXT
\vspace{-6mm}
\section{Introduction}
Humans rely crucially on cross-modal associations between sight and touch to physically interact with the world~\cite{smith2005development}. For example, our sense of sight tells us how the ground in front of us will feel when we place our feet on it, while our sense of touch conveys the likely visual appearance of an unseen object from a brief contact. Translating between these modalities requires an understanding of physical and material properties. Models trained to solve this problem must learn, for instance, to associate rapid changes in shading with rough microgeometry, and smooth textures with soft surfaces. 

Touch is arguably the most important sensory modality for humans~\cite{o1989sense,manske1999sense,linden2016touch}, due to its role in basic survival~\cite{linden2016touch,cox2006scn9a,hutmacher2019there} and physical interaction. Yet touch sensing has received comparably little attention in multimodal learning. An emerging line of work has addressed the problem of translating touch to sight, such as by learning joint embeddings~\cite{yang2022touch,lin2019learning}, manipulating visual styles to match a tactile signal~\cite{yang2022touch}, or adding a plausible imagery of a robotic arm to an existing photo of a scene~\cite{Li2019ConnectingTA}. While these tasks each capture important parts of the cross-modal prediction problem, each currently requires a separate, special-purpose method. Existing methods also lag significantly behind those of other areas of multimodal perception, which provide general-purpose methods for cross-modal synthesis, and can translate between modalities without the aid of extra conditional information. 

In this paper, we generate plausible images of natural scenes from touch (and vice versa), drawing on recent advances in diffusion models~\cite{ldm, dhariwal2021diffusion, Ho2020DenoisingDP, Ho2021CascadedDM, Nichol2021ImprovedDD}.  
We adapt latent diffusion models to a variety of visuo-tactile synthesis problems. 
Our proposed framework obtains strong results on several novel synthesis problems, and unifies many previously studied visuo-tactile synthesis tasks.

First, we study the problem of generating images from touch (and vice versa). We address the task of generating images from touch without any image-based conditioning, where we are the first method to successfully generate images for natural scenes (Fig.~\ref{fig:teaser}a).  We also address the task of adding an arm to a photo of an existing scene, where we significantly outperform prior work~\cite{Li2019ConnectingTA}. 

Second, we address the recently proposed {\em tactile-driven image stylization} task, \ie, the problem of manipulating an image to match a given touch signal~\cite{yang2022touch} (Fig.~\ref{fig:teaser}b), using an approach based on guided image synthesis~\cite{meng2022sdedit}. Our approach obtains results that are higher fidelity and that match the tactile signal significantly more closely than those of prior work. It also provides the ability to control the amount of image content preserved from the input image.

Finally, we show that we can augment our model with additional conditional information. Taking inspiration from the classic problem of intrinsic image decomposition~\cite{liu2020unsupervised, bell14intrinsic}, we perform {\em {tactile-driven shading estimation}}, predicting an image after conditioning on reflectance and touch (Fig.~\ref{fig:teaser}c). Since changes in tactile microgeometry often manifest as changes in shading (\ie, the information missing from reflectance), this tests the model's ability to link the two signals. We also use segmentation masks to create ``hand-less'' images that contain the object being pressed but not the tactile sensor or arm that pressed it.

We demonstrate our framework's effectiveness using natural scenes from the {\em Touch and Go} dataset~\cite{yang2022touch}, a collection of egocentric videos that capture a wide variety of materials and objects using GelSight~\cite{johnson2009retrographic}, and using robot-collected data from {\em VisGel}~\cite{Li2019ConnectingTA}. 

%-------------------------------------------------------------------------
\mysection{Related Work}
\paragraph{Cross-modal synthesis with diffusion models.}
Diffusion models have recently become a favored generative model family due to their ability to produce high-quality samples. However, one major concern for diffusion models is their slow inference speed due to the iterative generation process on high dimensional data. Recently, latent diffusion~\cite{ldm} addressed this drawback by working on a compressed latent space of lower dimensionality, which allows diffusion models to work on more extensive tasks with accelerating the speed. These models have demonstrated remarkable success in tasks such as image synthesis~\cite{dhariwal2021diffusion, Ho2020DenoisingDP, Ho2021CascadedDM, Nichol2021ImprovedDD}, super-resolution~\cite{Saharia2021ImageSV}, and image editing~\cite{Sinha2021D2CDM, meng2022sdedit, cheng2023wacv}. Additionally, the advancements in multimodal learning~\cite{ji2022mrtnet, ji2023online, Gao_2023_CVPR} have enabled diffusion models to be utilized for cross-modal synthesis tasks. For vision-language generation, diffusion models have been studied for text-to-image synthesis~\cite{Avrahami2021BlendedDF, Kawar2022ImagicTR, Nichol2021GLIDETP, Rahman2022MakeAStoryVM, Saharia2022PhotorealisticTD}, text-to-speech generation~\cite{Chen2022ResGradRD, Kong2020DiffWaveAV, Leng2022BinauralGradAT}, text-to-3D generation~\cite{luo2021diffusion, shao2022diffustereo}. In addition, diffusion models also show promising results in audio synthesis including text-to-audio generation~\cite{Singer2022MakeAVideoTG}, waveform generation~\cite{lam2022bddm, Lee2021PriorGradIC, Chen2022InfergradID}. In this work, we are the first to employ diffusion model on real-world visual-tactile data, exploring the possibility of utilizing tactile data as a prompt for image synthesis. In concurrent work, Higuera \etal~\cite{higuera2023learning} used diffusion to simulate tactile data, which they used to train a braille classifier.

\vspace{-1mm}
\mypar{Tactile sensing.}
Early touch sensors recorded simple, low-dimensional sensory signals, such as measures of force, vibration, and temperature~\cite{Lederman1987HandMA, Lederman2009TUTORIALRH, Cutkosky2008ForceAT}. Beginning with GelSight~\cite{Yuan2017GelSightHR,johnson2009retrographic}, researchers proposed a variety of vision-based tactile sensors, which convert the deformation of an illuminated membrane using a camera, thereby providing detailed information about shape and material properties~\cite{Taylor2021GelSlim3H, Lepora2022DigiTacAD}. We focus on these sensors, particularly using GelSight, since it is widely used applications~\cite{Li2019ConnectingTA, feel_success}, and available in visuo-tactile datasets~\cite{gao2021ObjectFolder, gao2022ObjectFolderV2, yang2022touch}. Crucially, these sensors produce images as output, allowing us to use the same network architectures for both images and touch~\cite{yuan2017shape}. Other work proposes collocated vision and touch sensors~\cite{yamaguchi2017implementing,chaudhury2022using}. 

\mypar{Cross-modal models for vision and touch.}
Li \etal~\cite{Li2019ConnectingTA} used a GAN~\cite{pix2pix2017} to translate between tactile signals and images, using a dataset acquired by a robot. 
 In contrast, they require conditioning their touch-to-image model on another photo from the same scene. This is a task that amounts to adding an arm grasping the correct object (given several possible choices), rather than generating an object that could have plausibly led to a touch signal according to its physical properties. It is not straightforward to adapt their method to the other touch-to-image synthesis problems we address without major modifications. Yang \etal~\cite{yang2022touch} proposed a visuo-tactile dataset and used a GAN to restyle images to match a touch signal. Their approach only learns a limited number of visual styles, and cannot be straightforwardly adopt extra conditional information (such as reflectance) or be applied to unconditional cross-modal translation tasks. Other work has learned multimodal visuo-tactile embeddings~\cite{yang2022touch,lin2019learning}. 
 % While we too learn visuo-tactile embeddings, we use them as part of a cross-modal synthesis model. 
 Other work learns to associate touch and sight for servoing and manipulation~\cite{chaudhury2022using}.

%-------------------------------------------------------------------------
\mysection{Method}
\begin{figure*}
    \centering
    \raggedright
    \vspace{-5mm} %
    \centering
    \includegraphics[width=0.95\textwidth]{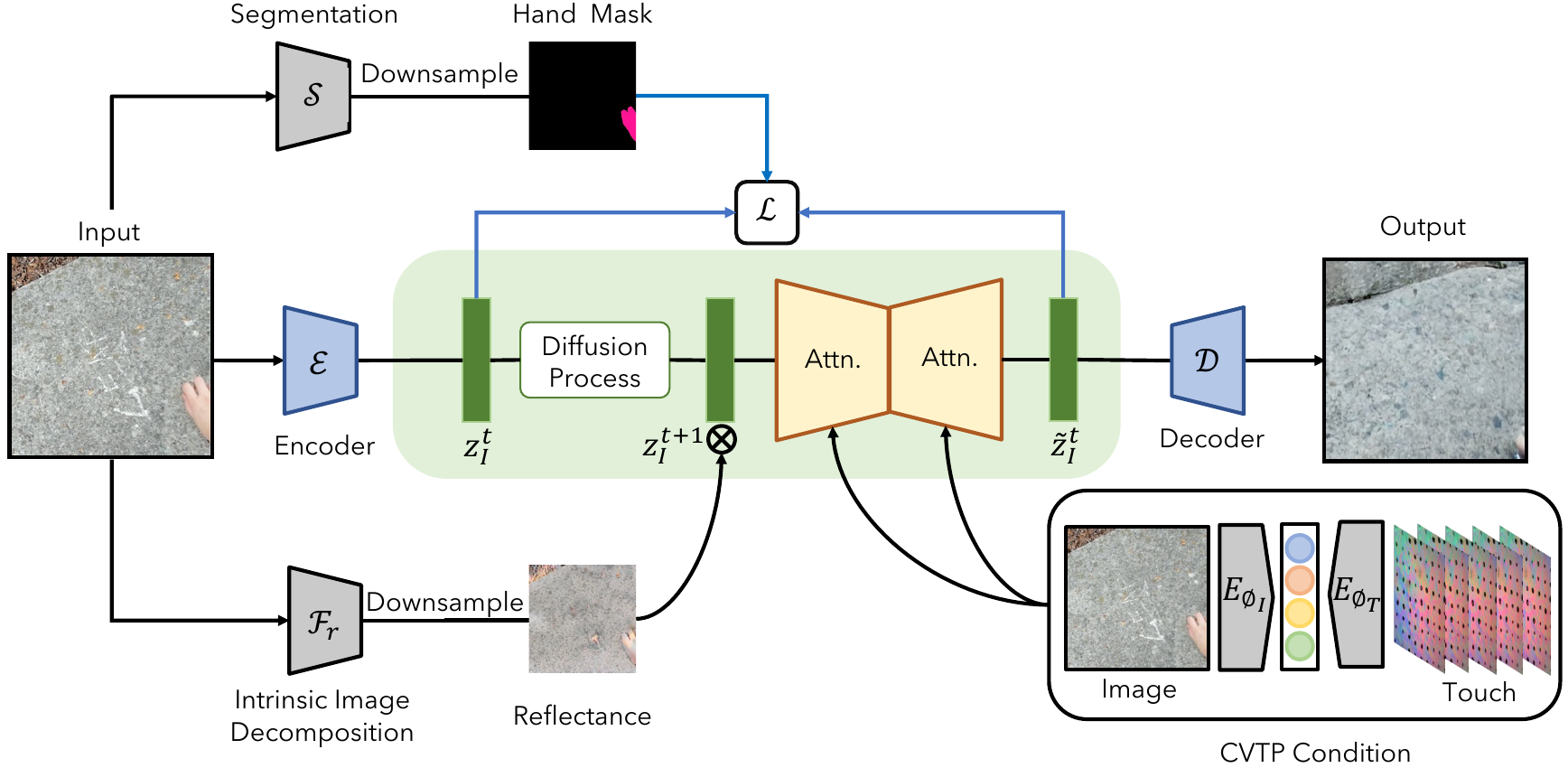}
    \vspace{-3mm}
    \caption{{\bf Touch-to-image model}. We use a latent diffusion model to generate an image of a scene from touch. The touch signal is represented using multiple frames of video from a GelSight sensor. The model uses a segmentation mask to optionally generate only the scene content containing the pressed object (\ie, without a hand or touch sensor). We also optionally condition on reflectance from a scene, in which case the model's generation task requires it to estimate shading.} \vspace{-4mm}
    \label{fig:pipeline}
\end{figure*}

Our goal is to translate touch to vision (and vision to touch) using a generative model. We will do this using a model based on latent diffusion~\cite{ldm}. We will use this model to solve a number of tasks, including: 1) cross-modal visual-tactile synthesis, 2) tactile-driven image stylization, and 3) tactile-driven shading estimation. 
\subsection{Cross-Modal Synthesis of Vision and Touch}
We now describe our framework for cross-modal synthesis. First, we describe a contrastive visuo-tactile model, which we use to perform conditional generation. Second, we describe our cross-modal latent diffusion model. 
 
\mysubsubsection{Contrastive Visuo-tactile Pretraining (CVTP)}\label{cvtp}

Following other work in cross-modal synthesis~\cite{radford2021learning,ldm}, we provide conditional information to our generation models through multimodal embeddings via contrastive learning~\cite{feng2023self, zheng2023exif, yang2022sparse, tian2020contrastive}. Our embedding-learning approach resembles that of Yang \etal~\cite{yang2022touch} and contrastive multiview coding~\cite{tian2020contrastive}. A key difference is that we incorporate temporal information into our visual and tactile representations. Touching an object is a dynamic process, and the information we obtain varies over time, from the moment when the tactile sensor begins touching the object, to the point when the sensor has reached it maximum deformation. Adding temporal cues provides information about material properties that may be hard to perceive from a single sample, such as the hardness or softness of a surface~\cite{yuan2017shape,ji2019human}.

%, and finally, to the moment when the tactile sensor leaves the object completely
%, which will help to distinguish between objects and increase the generation quality.

 Given the visual and tactile datasets $X_I$ and $X_T$, which consist of $N$ synchronized visual-tactile frames $\{\mathbf{x}_I^i,  \mathbf{x}_T^i\}_{i=1}^N$, we denote the video clip sampled at time $i$ with the window size $w=2C+1$, $v_I^i = \{\mathbf{x}_I^{i-C}, ...,  \mathbf{x}_I^{i}, ..., \mathbf{x}_I^{i+C}\}$ and the corresponding tactile clip $v_I^t = \{\mathbf{x}_I^{i-C}, ...,  \mathbf{x}_I^{i}, ..., \mathbf{x}_I^{i+C}\}$. We denote examples taken from the same visual-tactile recording $\{v_I^i,  v_T^i\}$ as positives, and samples from different visual-tactile video pair $\{\upsilon_I^i,  \upsilon_T^j\}$ as negatives.
 
Our goal is to jointly learn temporal visual $z_I = E_{\phi_I}(v_I)$ and tactile  $z_T = E_{\phi_T}(v_T)$ encoder. {We use a 2D ResNet as the architecture for both encoders. For easy comparison to static models, we incorporate temporal information into the model via early fusion (concatenating channel-wise).}

Then we maximize the probability of finding the corresponding visuo-tactile video pair in a memory bank containing $K$ samples using InfoNCE~\cite{oord2018representation} loss:
\begin{equation}\label{eq1}
  \mathcal{L}_{i}^{V_I, V_T} = -{\log}\frac{\exp(E_{\phi_I}(v_{I}^i) \cdot E_{\phi_T}(v_{T}^i)/\tau)}
  {\sum_{j = 1}^{K} {\exp}(E_{\phi_I}(v_{I}^i) \cdot E_{\phi_T}(v_{T}^j)/\tau)}
\end{equation}
where $\tau$ is a small constant. Analogously, we get a symmetric objective $\mathcal{L}^{V_T, V_I}$ and minimize: 
\begin{equation}\label{eq2}
  \mathcal{L}_{\text{CVTP}}= \mathcal{L}^{V_I, V_T} + \mathcal{L}^{V_T, V_I}.
\end{equation}

\vspace{-4mm}
\mysubsubsection{Touch-conditioned Image Generation} \label{ldm}
We now describe the tactile-to-image generation model (an image-to-touch model can be formulated in an analogous way). Our approach follows Rombach~\etal~\cite{ldm}, which translates language to images, but  with a variety of extensions specific to the visuo-tactile synthesis problem. Given a visuo-tactile image pair $\{\mathbf{x}_I,  \mathbf{x}_T\} \in \mathbb{R}^{H \times W \times 3} $, {our goal is to generate an image $\widetilde{\mathbf{x}}_I$ from tactile input $\mathbf{x}_T$}. 
We encode the input $\mathbf{x}$ into a latent representation $\mathbf{z} = \mathcal{E}(\mathbf{x}) \in \mathbb{R}^{h \times w \times 3}$. A decoder $\mathcal{D}$ will reconstruct the image $\hat{x} = \mathcal{D}(\mathbf{z})$ from the code. The latent dimension $h \times w$ is smaller than the image dimension $H \times W$. 

\mypar{Training.} We train a touch-to-vision diffusion generation in the latent space $\mathbf{z}_I = \mathcal{E}(\mathbf{x}_I)$. Diffusion models learn to generate images by recursively denoising from a normal distribution to the desired data distribution. Specifically, given our latent representation $\mathbf{z}_I$, we uniformly sample a diffusion step $t \in \{ 1, ..., T\}$ and 
obtain the corresponding noisy image $\mathbf{z}_{I}^{t}$ by iteratively adding Gaussian noise with a variance schedule. We use a U-Net~\cite{ronneberger2015u} network $\epsilon_\theta$ as our denoising model, which is conditioned on the tactile representation encoded through the tactile encoder $E_{\phi_T}$ trained in Section~\ref{cvtp}. We minimize:
\begin{equation} \label{eq:denoise}
    \begin{aligned}
        L(\theta, \phi) = \mathbb{E}_{\mathbf{z}_I, \mathbf{c}, \epsilon, t} \left[ \| \epsilon_t - \epsilon_\theta(\mathbf{z}_I^t, t, E_{\phi_T}(\mathbf{v}_T)) \|^2_2 \right],
    \end{aligned}
\end{equation}
where $\epsilon_t$ is the added noise at time $t$, and $\mathbf{v}_T$ is the  tactile example. The denoising network $\epsilon_\theta$ and the tactile encoder $E_{\phi_T}$ are jointly trained.

 %centered at $\mathbf{x}_T$ within the window size $w$
 
\mypar{Inference.} At test time, we first sample noise $\widetilde{\mathbf{z}}_I^T \sim \mathcal{N}(0,1)$ at time $T$, and then use the trained diffusion model to iteratively predict the noise $\widetilde{\epsilon}_t$, resulting in a denoised latent representation $\widetilde{\mathbf{z}}_I^t = \widetilde{\mathbf{z}}_I^{t+1} - \widetilde{\epsilon}_{t+1}$ from $t \in \{T - 1, ..., 0 \}$. Following~\cite{ldm, dhariwal2021diffusion}, we use classifier-free guidance to trade off between sample quality and diversity in the conditional generation, computing the noise as: %In formal terms:
\begin{equation}
    \begin{aligned}
        \widetilde{\epsilon}_t =  \epsilon_\theta(\widetilde{\mathbf{z}}_I^t, t, \emptyset) + s \cdot \left(\epsilon_\theta(\widetilde{\mathbf{z}}_I^t, t, E_{\phi_T}(\mathbf{v}_T)) - \epsilon_\theta(\widetilde{\mathbf{z}}_I^t, t, \emptyset)\right),
    \end{aligned}
\end{equation}
where $\emptyset$ denotes a zero-filled conditional example (for unconditional generation), and $s$ is the guidance scale. Finally, we convert the latent representation $\widetilde{\mathbf{z}}_I^0$ to an image $\widetilde{\mathbf{x}}_I = \mathcal{D}(\widetilde{\mathbf{z}}_I^0) \in \mathbb{R}^{H \times W \times 3}$.

\begin{figure}[t]
  \centering
  \footnotesize
  \setlength{\abovecaptionskip}{0.1cm}
  \includegraphics[width=1\linewidth]{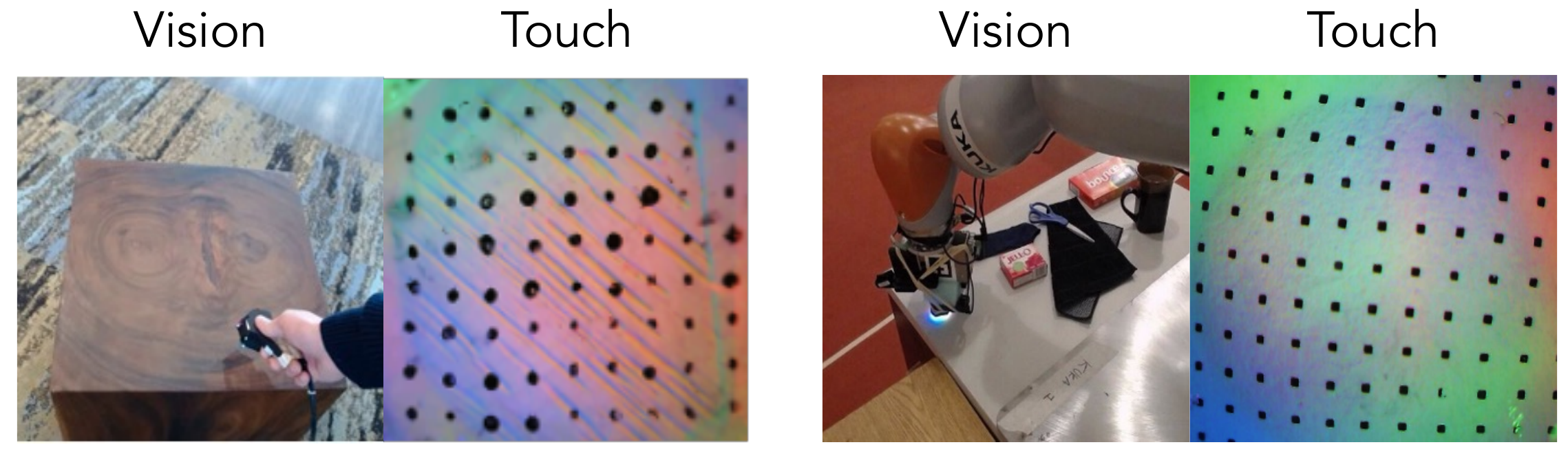}
    \begin{flushleft}
    \small
    \vspace{-2mm}
    \hspace{8mm} \small{Touch and Go~\cite{yang2022touch}\hspace{23mm} VisGel~\cite{Li2019ConnectingTA}}
    \end{flushleft}
  
  \vspace{-0.5mm} \caption{{\bf Visuo-tactile datasets}. For our experiments, we evaluate our model on natural scenes from {\em Touch and Go}~\cite{yang2022touch} and robot-collected data from {\em VisGel}~\cite{Li2019ConnectingTA}. \vspace{-.5mm}}
  \label{fig:dataset_comparison}
  \vspace{-1em} 
\end{figure}

\mysubsection{Visuo-Tactile Synthesis Models}

So far, we have presented models for translating between touch and images (and vice versa). We now describe several visuo-tactile synthesis models that we build on this diffusion framework.

\mysubsubsection{Generating realistic images without hands}

One of the challenges of dealing with visuo-tactile data is that the tactile sensor typically occludes the object that is being touched (Fig.~\ref{fig:dataset_comparison}). Generated images will therefore contain the sensor, and potentially the arm that held it. This is not always desirable, as a major goal of touch sensing is to generate images of objects or materials that could have plausibly led to a given touch signal. We address this problem for the natural scenes from the {\em Touch and Go} dataset~\cite{yang2022touch}, which contain visible human hands and GelSight sensors~\cite{Yuan2017GelSightHR}.

To generate images containing only objects that yield a given tactile signal (without hands or touch sensors), we only compute the loss for pixels that do not overlap with hands during the training, thereby depriving the model of supervision for hand pixels. We first generate hand segmentation masks for the visual image $\mathbf{m}_I = \mathcal{S}(\mathbf{x}_I)$ and obtain the downsampled mask $\mathbf{z}_m$ of the same spatial dimension of the image latent representation. {For this, we use the off-the-shelf hand segmentation model from Darkhalil et al.~\cite{VISOR2022}, which is a modified model from PointRend~\cite{Kirillov2019PointRendIS} instance segmentation designed specifically for segmenting hands.}
We then mask the diffusion loss (Eq.~\ref{eq:denoise}) to be: 
\begin{equation}
    \begin{aligned}
        %L(\theta, \phi) =
        \mathbb{E}_{\mathbf{z}_m, \mathbf{z}_I, \mathbf{c}, \epsilon, t} \left[ \| \mathbf{z}_m \odot \left(\epsilon_t - \epsilon_\theta(\mathbf{z}_I^t, t, E_{\phi_T}(\mathbf{v}_T)) \right)\|^2_2 \right],
    \end{aligned}
\end{equation}
where $\mathbf{z}_m$  indicates whether a pixel overlaps with a hand, and $\odot$ denotes pointwise multiplication. 

\mysubsubsection{Tactile-driven Image Stylization}\label{tdis}
Tactile-driven image stylization~\cite{yang2022touch} aims to manipulate the visual appearance of an object so that it looks more consistent with a given touch signal. Previous work posed the problem of editing the visual style of an image while preserving its structure~\cite{yang2022touch,li2021learning}.

Given an input image $\mathbf{x}_I$ and a desired tactile signal $\mathbf{x}_T'$ (obtained from a different scene), our goal is to manipulate $\mathbf{x}_I$ so that it appears to ``feels'' more like  $\mathbf{x}_T'$. We adapt the approach of Meng \etal~\cite{meng2022sdedit}. We first compute the noisy latent representation $z_I^{N}$ at time $0 \leq N \leq T$, where $T$ denotes the total number of denoising steps.  We then conduct the denoising process for $z_I^{N}$ from time step $N$ to 0 conditioned on $\mathbf{x}_T'$. This allows for fine-grained control over the amount of content preserved from the input image, via the parameter $N$. We analyze the choice of $N$ at Sec.~\ref{control}.

\begin{figure*}[t]
\begin{center}
   \includegraphics[width=1.0\linewidth]{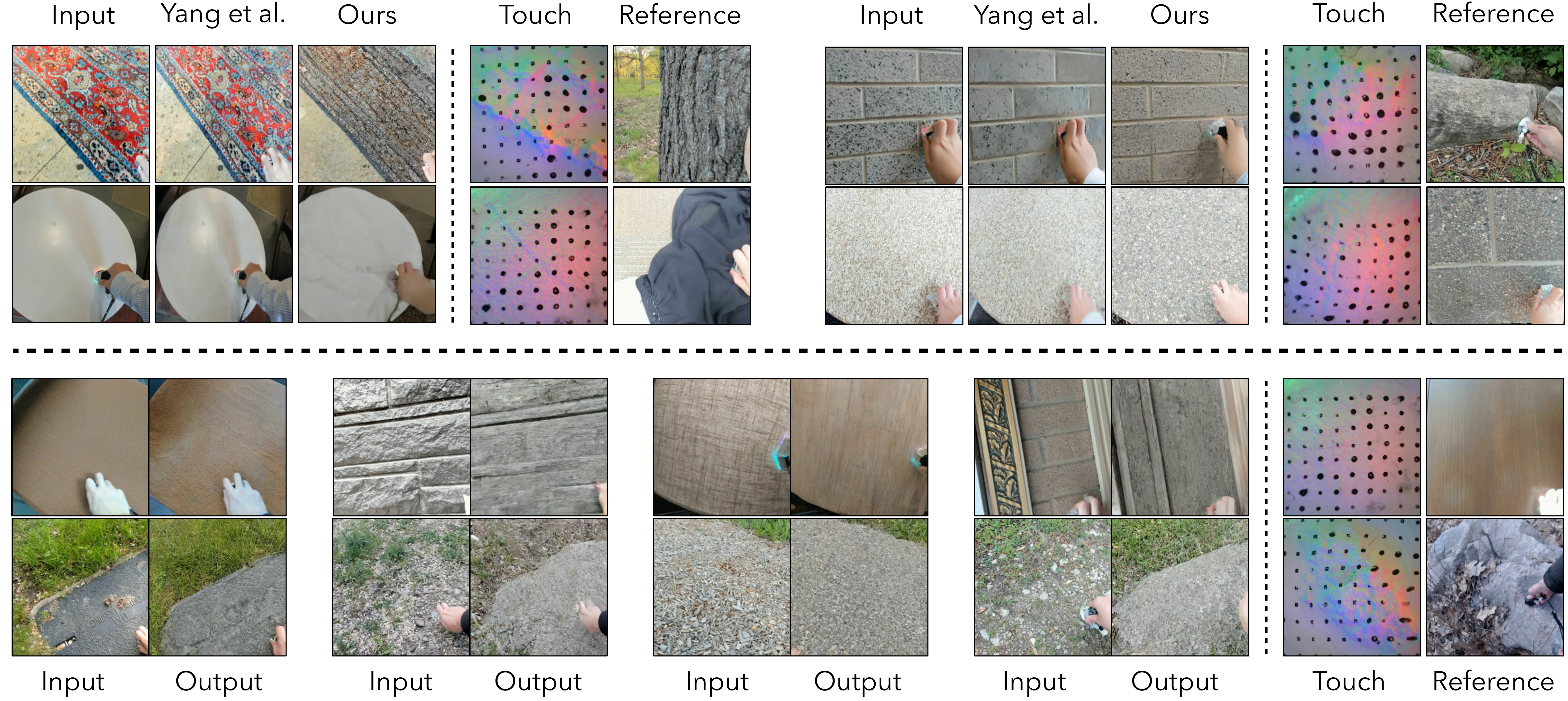}
\end{center}
\vspace{-5mm}   
\caption{\textbf{Tactile-driven Image Stylization.} \emph{(Top)} We restyle the input image using the given touch signal (reference image from scene provided for clarity). We compare our approach to Yang et al.~\cite{yang2022touch}. Our approach generates images with higher quality matching more closely to the given tactile signal. \emph{(Bottom)} We show more examples of the manipulated images. Please see supplement for more examples.}
\label{fig:tdis}
\vspace{-3mm}
\end{figure*}
\mysubsubsection{Tactile-driven Shading Estimation}

Touch conveys a great deal of information about a surface's microgeometry~\cite{johnson2009retrographic}. Much of this information can also be perceived through {\em shading} cues: intensity variations due to light interacting with surface orientation for objects with Lambertian material properties. Following classic work in intrinsic image decomposition~\cite{barrow1978recovering,grosse2009ground,bell14intrinsic}, we assume that the image can be factorized into reflectance and shading for each pixel, i.e., we can write our image $\mathbf{x}_I = \mathbf{x}_R \odot \mathbf{x}_S$ where the two terms in the product are the per-pixel reflectance and shading. 

We propose a model that deals with inferring shading from touch.  Given an image's estimated reflectance map $\mathbf{x}_R$, along with a touch signal $\mathbf{x}_T$, we reconstruct the original image $\mathbf{x}_I$. This is a task that requires inferring the shading, since it is the component that is missing from the input. By formulating the problem so that we predict the original image, we can easily reuse the latent encoder/decoder from natural images.

{We address this task by modifying our network so that it also takes reflectance as input (Eq.~\ref{eq:denoise}). We first estimate reflectance using the intrinsic image decomposition model of Liu~\etal~\cite{liu2020unsupervised} and downsample it to the same dimensions as the latent space. We then concatenate the downsampled reflectance $\mathbf{z}_R$ to the noisy representation $\mathbf{z}_I^t$ as the input for each denoising step. Thus we modify the loss function (Eq.~\ref{eq:denoise}) as the following:}
\begin{equation} \label{eq:denoise}
\small
    \begin{aligned}
        L(\theta, \phi) = \mathbb{E}_{\mathbf{z}_I, \mathbf{c}, \epsilon, t} \left[ \| \epsilon_t - \epsilon_\theta(\mathbf{z}_I^t \otimes \mathbf{z}_R, t, E_{\phi_T}(\mathbf{v}_T)) \|^2_2 \right],
    \end{aligned}
\end{equation}
where $\otimes$ denotes concatenation.

%------------------------------------------------------------------------
\mysection{Results}
We evaluate our cross-modal synthesis models through qualitative and quantitative experiments on natural scenes and robot-collected data. 

\mysubsection{Implementation details}\label{Implementation}
\paragraph{Contrastive visuo-tactile model.} Following~\cite{yang2022touch}, we use ResNet-18 as the backbone of contrastive model, and train on {\em Touch and Go}~\cite{yang2022touch}. This model is trained using SGD for 240 epochs with the learning rate of $0.1$ and weight decay of $10^{-4}$. The ResNet takes 5 reference frames as input using early fusion (concatenated channel-wise) and we take the feature embedding from the last layer of the feature and map it to 512 dimensions. Following prior work~\cite{tian2020contrastive}, we use $\tau=0.07$ and use a memory bank with 16,385 examples.

\mypar{Visuo-tactile diffusion model.} 
 We base our latent diffusion model on Stable Diffusion~\cite{ldm}. We use the Adam optimizer with the base learning rate of $2 \times 10^{-6}$. Models are all trained with 30 iterations using the above learning rate policy. We train our model with the batch size of 96 on 4 RTX A40 GPUs.  The conditional model is finetuned along with the diffusion model. We use the frozen, pretrained VQ-GAN~\cite{Ding2021VQGNNAU} to obtain our latent representation, with the spatial dimension of 64$\times$64. During the inference, we conduct denoising process for 200 steps and set the guidance scale $s=7.5$.
 \begin{figure*}[t]
\begin{center}
% \fbox{\rule{0pt}{2in} \rule{0.9\linewidth}{0pt}}
   \includegraphics[width=1.0\linewidth]{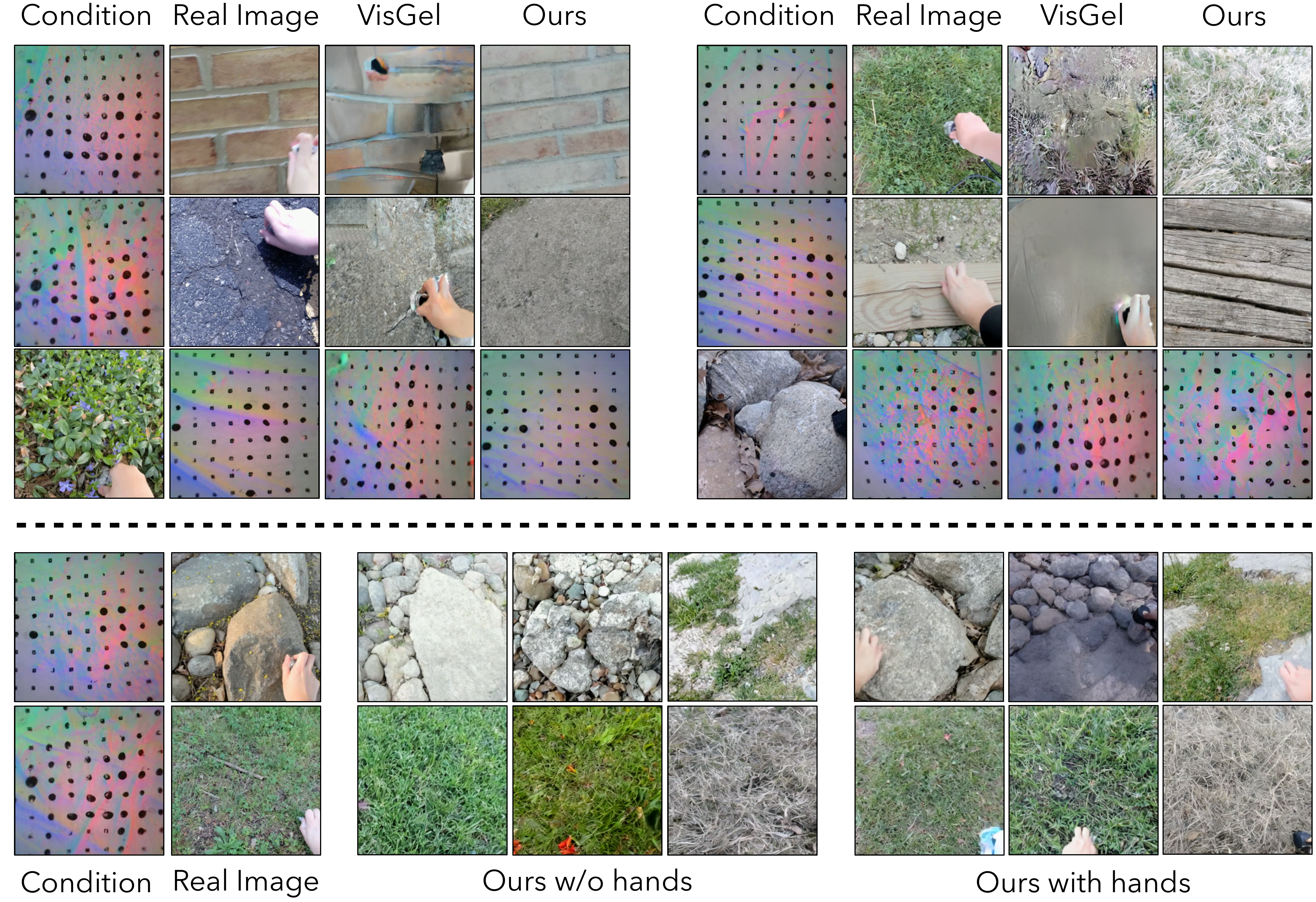}
\end{center}
\vspace{-5mm}
   \caption{\textbf{Visuo-tactile Cross Generation on \emph{Touch and Go} dataset.} \emph{(Top)} We compare our approach to state-of-the-art method Visgel~\cite{Li2019ConnectingTA}. \emph{(Bottom)} We show more results of our generated images with and without hands. In both case our approach is able to generate realsitic images with high fidelity.}
   \vspace{-5mm}
\label{fig:touchandgo}
\end{figure*}

\mysubsection{Experimental Setup}
\paragraph{Dataset.} We conduct our experiments on two real-world visuo-tactile datasets: 
\vspace{-2mm}
\begin{itemize}[leftmargin=10pt]
\setlength\itemsep{0mm}

    \item \textbf{\emph{Touch and Go} dataset.} The \emph{Touch and Go} dataset is a recent, real-world visuo-tactile dataset in which humans probe a variety of objects in both indoor and outdoor scenes. There are 13,900 touches from roughly 4000 different object instances and 20 material categories. Since this is the only available dataset with zoomed-in images and clearly visible materials, we use it for all three tasks. % including both indoor (52.2\%) and outdoor (47.8\%).
    \item \textbf{\emph{VisGel} dataset.} The \emph{VisGel} dataset contains synchronized videos of a robot arm equipped with a GelSight sensor interacting with 195 household objects. The dataset includes 195 objects from a wide range of indoor scenes of food items, tools, kitchen items, to fabrics and stationery. In total, the dataset contains 12k touches and around 3M frames.
\end{itemize}
\vspace{-2mm}

\begin{table}[t]
        \small
	\centering
	\caption{Evalutation of cross-modal generation on  \emph{Touch and Go}.}
        \vspace{-2mm} 
	\setlength{\tabcolsep}{0mm}{
	\begin{tabular}{@{}lcccccc@{}}
        \toprule
        \multicolumn{1}{l}{\multirow{2}{*}{\textbf{Method}}} & \multicolumn{3}{c}{\textbf{Touch $\rightarrow $ Image}} &  & \multicolumn{2}{c}{\textbf{Image $\rightarrow $ Touch}} \\ \cmidrule(l){2-7} 
        \multicolumn{1}{c}{} & CVTP ($\uparrow$) & Material($\uparrow$) & FID($\downarrow$) &  & SSIM($\uparrow$)  & PSNR($\uparrow$) \\ \midrule
        % Target &  0.13 & -  & - &  & - & - \\
        % \midrule
        Pix2Pix~\cite{pix2pix2017} & 0.08  &  0.15 & 136.4  &  &0.43  & 14.3 \\
        VisGel~\cite{Li2019ConnectingTA} &0.07  & 0.15 &128.3  &  &0.45  &15.0\\
        % \midrule
        
        Ours w/ hands &  \textbf{0.12} &  0.22 & \textbf{48.7} &  & \textbf{0.50}  & \textbf{15.4} \\
        Ours w/o hands &  \textbf{0.12}&  \textbf{0.24} & 81.5 &  & \textbf{0.50}  & \textbf{15.4}  \\
        \bottomrule
        \end{tabular}} \vspace{-3mm}
        \label{tab:touchandgo}
\end{table}

\mypar{Evaluation metrics.} We use several quantitative metrics to evaluate the quality of our generated images or tactile signals. We use {\bf Frechet Inception Distance (FID)},  which compares the distribution of real and generated image activations using trained network. Following Yang~\etal~\cite{yang2022touch} and CLIP~\cite{radford2021learning}, we take the cosine similarity between our learned visual and tactile embeddings for the generated images and conditioned tactile signals, a metric we call \textbf{Contrastive Visuo-Tactile Pre-Training (CVTP)}. A higher score indicates a better correlation between touch and images. It is worth noting that the CVTP metric only takes one frame of touch input. 
 Following~\cite{yang2022touch}, we measure \textbf{Material Classification Consistency}: we use the material classifier from Yang~\etal~\cite{yang2022touch} to categorize the predicted and ground truth images, and measure the rate at which they agree. 
    
Finally, following~\cite{Gao_2023_CVPR}, we evaluate standard \textbf{Structural Similarity Index Measure (SSIM)} and \textbf{Peak Signal to Noise Ratio (PSNR)}~\cite{Wang1987SSIMAS} metrics. 
\begin{figure*}[t]
\begin{center}
   \includegraphics[width=1.0\linewidth]{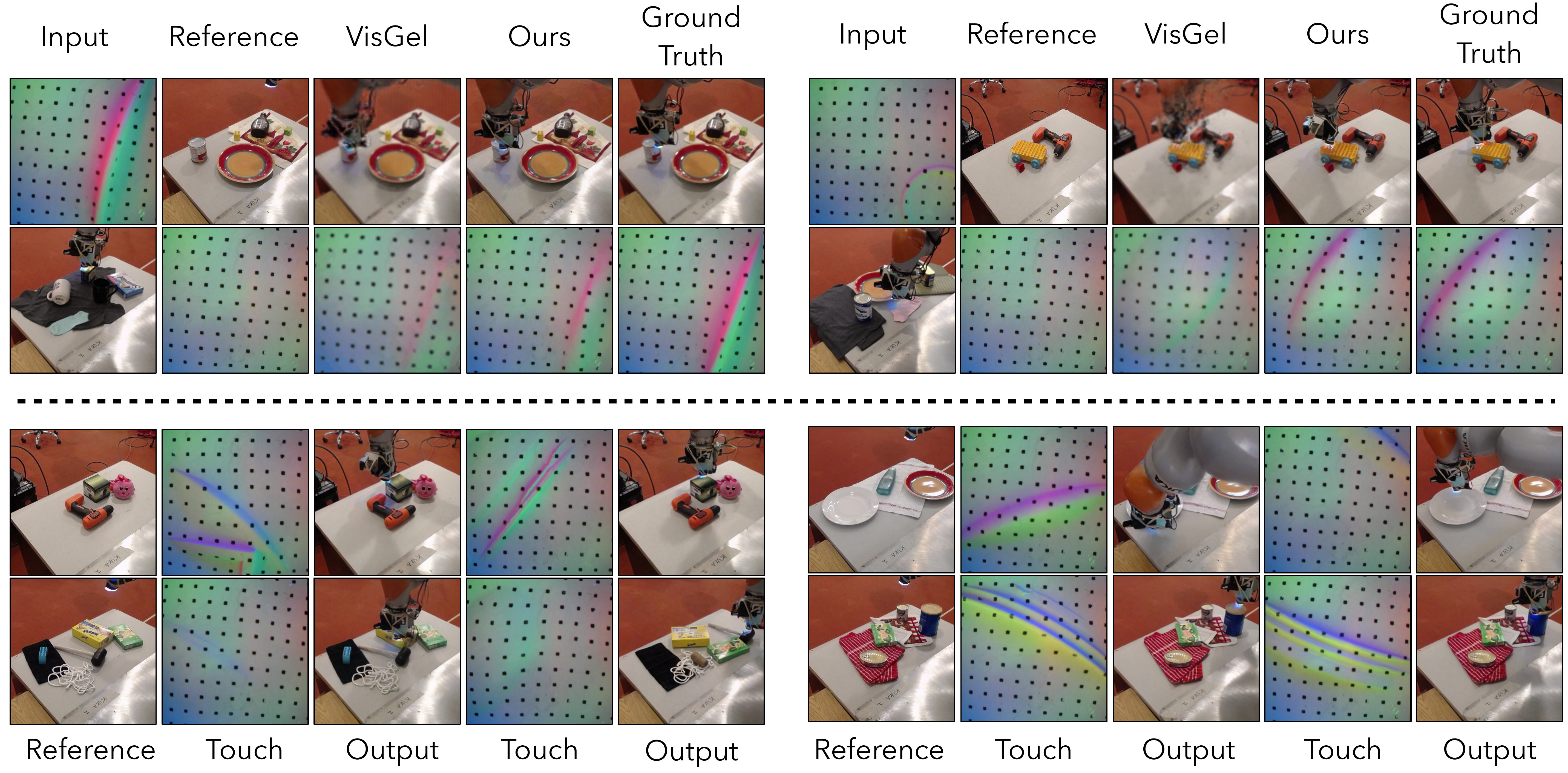}
\end{center}
\vspace{-5mm}   
   \caption{\textbf{Visuo-tactile Cross Generation on \emph{VisGel} dataset.} \emph{(Top)} We compare our approach to state-of-the-art method VisGel~\cite{Li2019ConnectingTA}. \emph{(Bottom)} Our approach is able to generate robotic hands touching reasonable locations of objects given the same reference image but different tactile signals.}
   \vspace{-5mm}
\end{figure*}
\begin{table}[t]
        \small
	\centering
	\caption{Evaluation of cross-modal generation on \emph{VisGel} (and conditioning on another photo from the scene).}
        \vspace{-2mm}
	\setlength{\tabcolsep}{1.2mm}{
	\begin{tabular}{@{}lccccc@{}}
        \toprule
        \multicolumn{1}{l}{\multirow{2}{*}{\textbf{Method}}} & \multicolumn{2}{c}{\textbf{Touch $\rightarrow $ Image}} &  & \multicolumn{2}{c}{\textbf{Image $\rightarrow $ Touch}} \\ \cmidrule(l){2-6} 
        \multicolumn{1}{c}{} & SSIM($\uparrow$) & PSNR($\uparrow$) &  & SSIM ($\uparrow$)  & PSNR ($\uparrow$) \\ \midrule
        % Target &  0.061 & -  &  & xxx & xxx \\
        % \midrule
        Pix2Pix~\cite{pix2pix2017} & 0.50  &15.1  &  &0.71  & 20.7 \\
        VisGel~\cite{Li2019ConnectingTA} &0.59  & 17.9  &  &0.76  &26.2\\
        % \midrule
        Ours &  \textbf{0.76} & \textbf{21.5} &  & \textbf{0.85} &  \textbf{27.6}  \\
        \bottomrule
        \end{tabular}}\vspace{-2mm}
        \label{tab:visgel}
        \vspace{-5mm}
\end{table}

\mysubsection{Cross-modal Generation}
We perform cross-modal generation, \ie, generating an image from touch and vice versa, on both in-the-wild \emph{Touch and Go} dataset and robot-collected dataset \emph{VisGel}. For straightforward comparison to prior work~\cite{Li2019ConnectingTA}, on \emph{VisGel} we provide a {\em reference} photo of the scene as an input to the model. Thus, successfully predicting the ground truth image amounts to inserting imagery of the robotic arm to the correct location in the scene. For {\em Touch and Go}, we do not condition the model on a visual input: instead, we simply translate one modality to the other. 

For evaluation metrics, we use CVTP, material classification consistency, and FID score for touch-to-image generation and SSIM and PSNR for image-to-touch generation. For \emph{VisGel} dataset we leverage SSIM and PSNR as the evaluation metric for both tasks. We only use CVTP, material classification consistency and FID only on touch-to-image generation task on \emph{Touch and Go}, since these evaluation metrics rely on a pretrained neural network from datasets of natural images, which may not generalize well on a different modality or to robot-collected data.

We compare our model to the prior state-of-the-art visuo-tactile generation method~\cite{Li2019ConnectingTA}, which is adapted from pix2pix~\cite{pix2pix2017} and is specifically designed to bridge the large domain gap between modalities by adding a reference image and temporal condition. As it is not possible to find a reference image in the natural image dataset, we remove the reference image while keeping everything else the same.

We show quantitative results for both tasks on \emph{Touch and Go} and \emph{VisGel} in Table~\ref{tab:touchandgo} and Table~\ref{tab:visgel} respectively. Our methods outperform existing state-of-the-art methods by a large margin for all evaluation metrics. We note that the variation of our model that removes hands from images obtains a worse FID score compared to those with hands, due to the discrepancy of hands between the original dataset and our generated images. Interestingly, the presence of hands does not does not affect the performance of CVTP and material classification consistency. We provide qualitative results from both models in Figure~\ref{fig:touchandgo} \emph{(bottom)}.

\begin{table}[t]
\small
        % \label{tab:tdis}
	\centering\vspace{1mm}  
	\caption{Quantitative results of of tactile-driven image stylization.}
    \vspace{-1mm}   
	\setlength{\tabcolsep}{2.5mm}{
	\begin{tabular}{l c c c}
		\toprule
    		\multirow{2}*{\textbf{Method}} & \multicolumn{3}{c}{\textbf{Evaluation Metrics}} \\ \cmidrule{2-4} & CVTP ($\uparrow$) & Material ($\uparrow$) & FID ($\downarrow$)\\
		\midrule
		Cycle GAN~\cite{UIIT_Cycle_Gan_Zhu} & 0.09 & 0.15 & 24.6\\
		Yang et al. \cite{yang2022touch} & 0.10 & 0.20 & 22.5\\
            Ours& \textbf{0.13} & \textbf{0.22} & \textbf{15.8}\\
		\bottomrule
	\end{tabular}}
 \vspace{-4mm}  
 \label{tab:tdis}
\end{table}

\mysubsection{Tactile-Driven Image Stylization}
Following~\cite{yang2022touch}, we evaluate the performance of tactile-driven image stylization on \emph{Touch and Go}~\cite{yang2022touch} using CVTP and material classification metrics. We also calculate the FID score between the set of generated images and the set of real images associated with the given tactile signals, which measures the fidelity of the output. We compare our model to a modified version of CycleGAN~\cite{UIIT_Cycle_Gan_Zhu} and the state-of-the-art method of Yang et al.~\cite{yang2022touch}. From the quantitative comparisons in Table~\ref{tab:tdis}, our method demonstrates a significant improvement over existing methods. We also show  qualitative comparisons in Figure~\ref{tab:tdis}, where the generated images more closely match the tactile signal, and we are able to generate styles that existing methods fail to capture. %fail to manipulate.

\begin{figure}[t]
\begin{center}
   \includegraphics[width=1.0\linewidth]{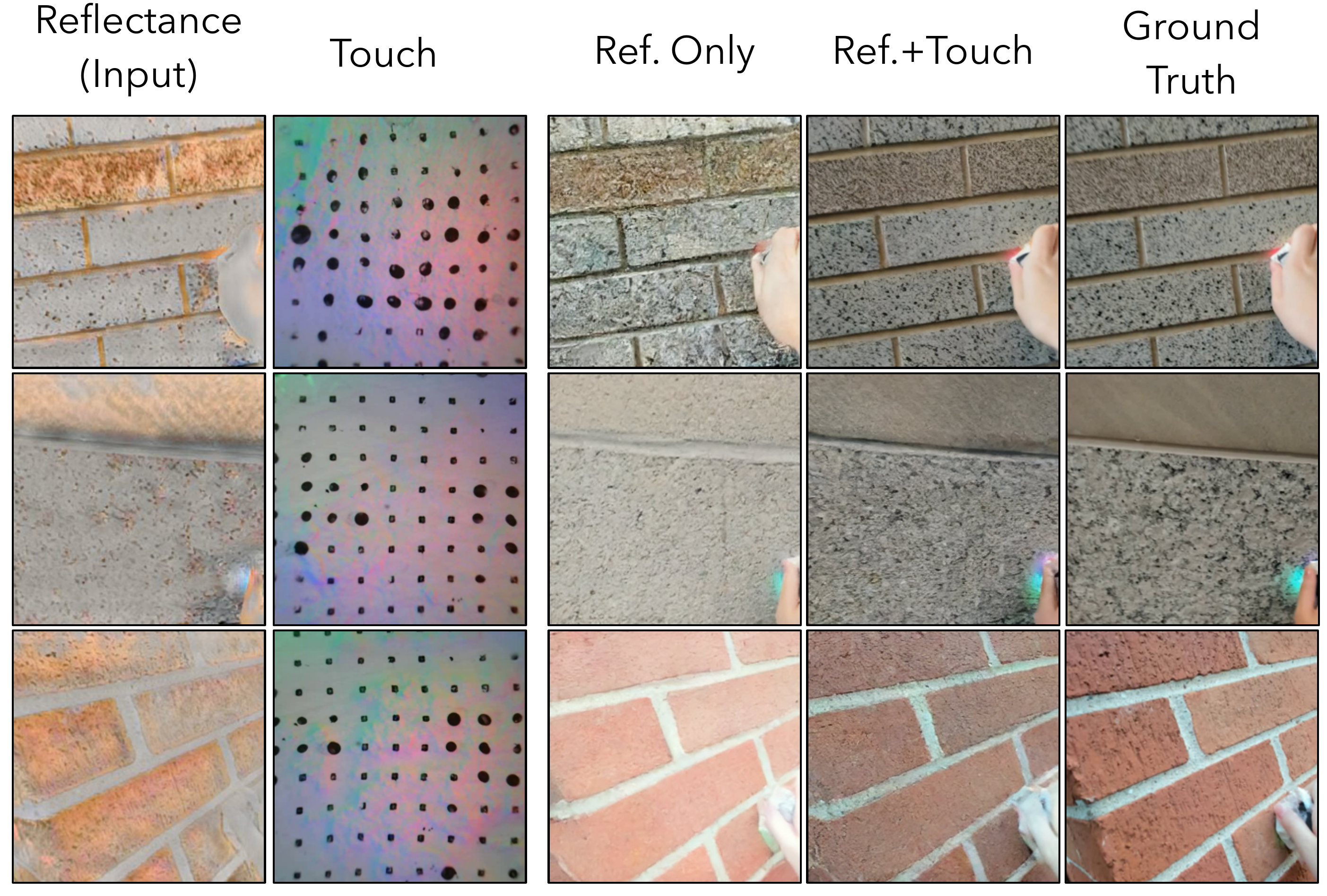}
\end{center}
\vspace{-5mm}   
\caption{\textbf{Tactile-driven shading estimation.} {We compare our approach to a model without a tactile signal (only reflectance), finding that the tactile-driven model better captures subtle material properties, such as roughness.}}
\label{fig:Ref2img}
\vspace{-3mm}
\end{figure}

\mysubsection{Tactile-driven Shading Estimation}
We hypothesize that the tactile signal conveys information about the microgeometry of an image, and thus allows a model to produce more accurate images than a reflectance-to-image model that does not have access to touch. We evaluated both models on {\em Touch and Go}  (Table~\ref{tab:ref2img}) and found that adding touch indeed improves performance on all evaluation metrics. We also show qualitative comparisons in Figure~\ref{fig:Ref2img}. We found that tactile signals are especially informative for predicting roughness and smoothness of Lambertian surfaces, such as bricks.

\begin{table}[t]
\small
	\centering
	\caption{Quantitative results for {tactile-driven shading estimation}.}
        \vspace{-2mm}   
	\setlength{\tabcolsep}{2.5mm}{
	\begin{tabular}{@{}lcccc@{}}
        \toprule
        \multicolumn{1}{l}{\multirow{2}{*}{\textbf{Method}}} & \multicolumn{3}{c}{\textbf{Reflectance $\rightarrow $ Image}} &  \\ \cmidrule(l){2-4} 
        \multicolumn{1}{c}{} & SSIM($\uparrow$) & PSNR($\uparrow$) & FID($\downarrow$) \\ \midrule
        Touch Only & 0.27  &  11.6 & 48.7\\
        Reflectance Only & 0.46  &  14.5 & 40.7\\
        Reflectance + Touch & \textbf{0.48}  & \textbf{15.4} & \textbf{36.9} \\
        \bottomrule
        \end{tabular}}
        \vspace{-3mm}  
        \label{tab:ref2img}
\end{table}

\mysubsection{Analysis}

\begin{figure}[t]
\begin{center}
   \includegraphics[width=0.95\linewidth]{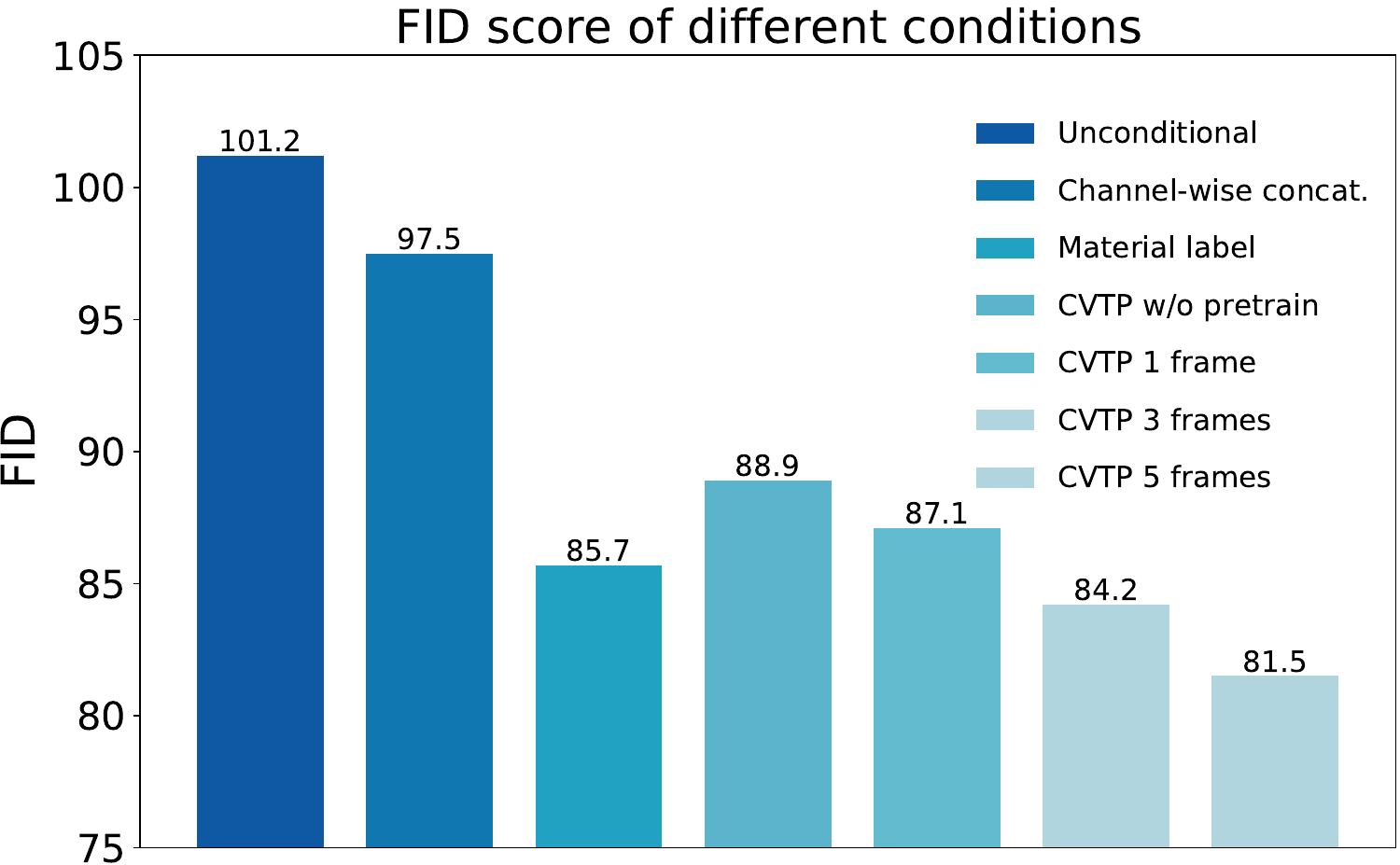}
\end{center}
\vspace{-5mm}   

\caption{{\bf Effect of different types of tactile conditioning}.}
\vspace{-2mm}
\label{fig:condition}
\end{figure}

\paragraph{Importance of temporal information.}
We first study the effect of adding multiple GelSight frames to the contrastive visuo-tactile embedding (Figure~\ref{fig:condition}). We compare our method with the unconditional generation and material class conditional generation on \emph{Touch and Go}. We found that conditioned generation provides a large improvement in performance compared to the unconditional generation. We also observed that the generation conditioned on the pretrained model is significantly better than that without pretraining. Interestingly, the model conditioned on the material class outperforms the variation of the model that only observes a single GelSight frame, suggesting that perceiving a touch signal from only a single moment in time may be less informative than the material category. Providing the model with additional frames significantly improves the model, with the 5-frame model obtaining the overall best performance. 

\mypar{Controllable Image Stylization}\label{control}
\begin{figure}%[t]
\begin{center}
   \includegraphics[width=0.85\linewidth]{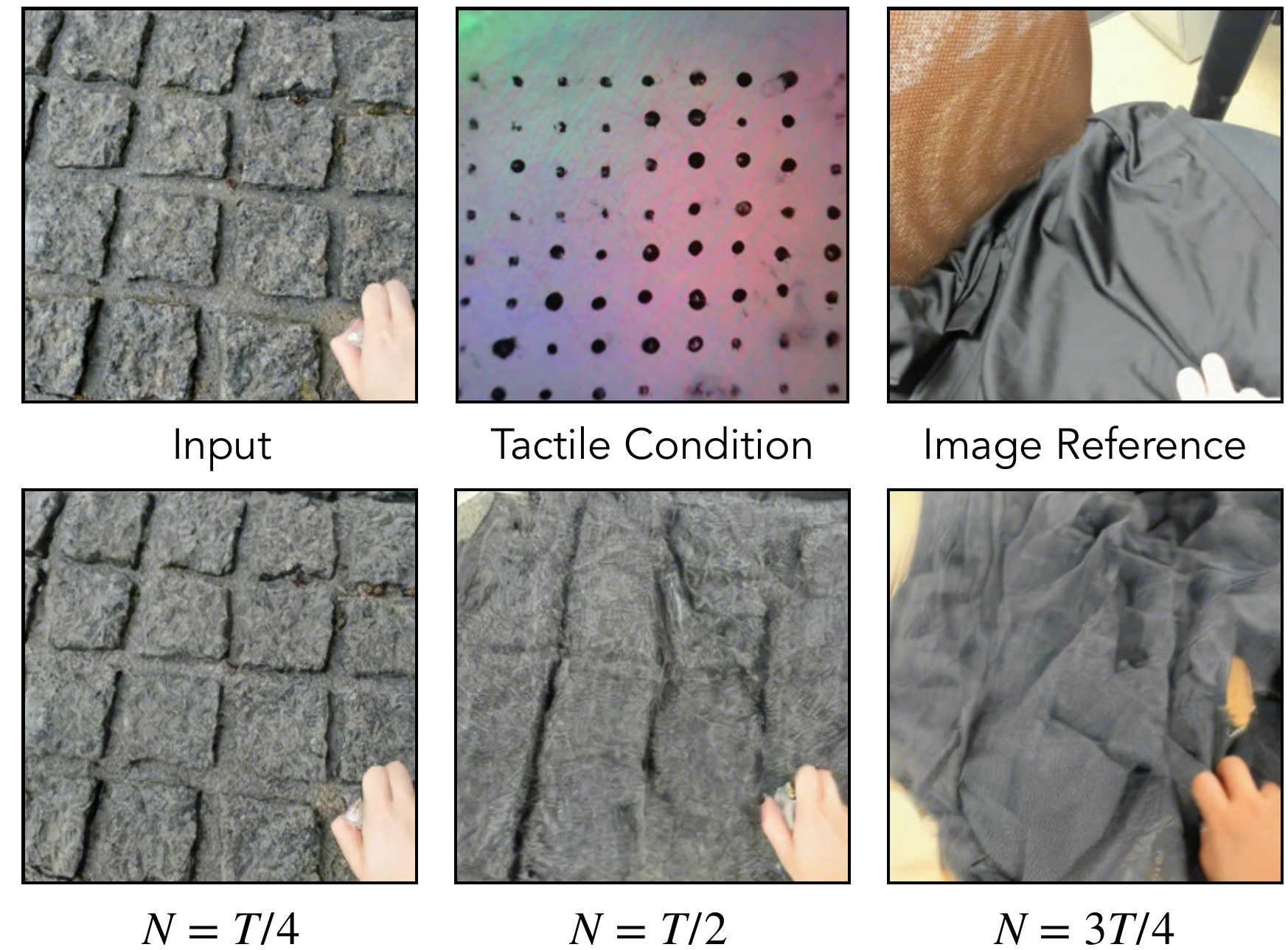}
\end{center}
\vspace{-4mm}   
\caption{{\bf Controlling the amount of preserved image content. } Manipulated images of tactile-driven image stylization using different values of $N$.}\vspace{-4mm}
\label{fig:tdis_n0}
\end{figure}
 
Our method allows us to control over the amount
of image content that is preserved from the original image by changing the denoising staring point $N$ (Sec.~\ref{tdis})~\cite{meng2022sdedit}. From Figure~\ref{fig:tdis_n0}, we observe that if we select the larger $N$, the generated image will be changed more drastically where the visual appearance will be changed completely to match the tactile signal while ruining the original image structure. In extreme case, where $N=T$ the manipulated result will be equal to the touch-to-image generation result, while small $N$ will result in little overall change. We empirically found that selecting $N=T/2$ obtains a good trade-off between these factors. 

%-------------------------------------------------------------------------
\mysection{Conclusion}
We proposed a visuo-tactile diffusion model that unifies previous cross-modal synthesis tasks, and allows us to address novel problems. We are the first to generate realistic images in the natural scenes from touch (and vise versa) without any image-based conditioning. We also show the ability to generate realistic ``hand-less'' images and solve a novel {tactile-driven shading estimation} task. 
Finally, we obtain significantly more realistic results on the tactile-driven stylization task than prior work. We see our work as being a step toward integrating the fields of tactile sensing and generative modeling.

\vspace{-1mm}
\mypar{Limitations.} Since our work has applications in creating fake imagery, a potential issue is that it could be used to create disinformation. Also, as touch mainly conveys material properties and microgeometry, the generated image will often differ semantically from the ground truth. 
%-------------------------------------------------------------------------
% \newpage
\vspace{-1mm}
\mypar{Acknowledgements.} We thank Chao Feng, Ziyang Chen and Shaokai Wu for the helpful discussions and help for visualizations. This work was supported in part by Cisco Systems.

{\small
\bibliographystyle{ieee_fullname}
\bibliography{main}
}
\clearpage
\appendix
\onecolumn
We provide additional details about our method, and provide qualitative results for our generation tasks.

\section{Model Architecture and Implementation Details}

We provide additional details about the latent diffusion model, such as the training hyperparameters.

% Please add the following required packages to your document preamble:
% \usepackage{booktabs}
\begin{table*}[h]
\centering
\caption{We show detailed hyperparamters setting of our models, including first stage model, condition model and LDM model.}
\label{tab:my-table}
\begin{tabular}{@{}ll|ll@{}}
\toprule
Hyperparamter & Value & Hyperparamter & Value \\ \midrule
Learning Rate & $2 \times 10^{-6}$ & LDM Model & U-Net \\
Image Size & 256 & LDM Input Size & 64 \\
Channel & 3 & LDM Input Channel & 3 \\
Conditioning Key & Crossattn & LDM Output Channel & 3 \\
First Stage Model & VQModelInterface & LDM Attention Resolutions & {[}8,4,2{]} \\
VQ In-channel & 3 & LDM Num Resblocks & 2 \\
VQ Out-channel & 3 & LDM Channel Mult & {[}1,2,3,5{]} \\
VQ Num. Resblocks & 2 & LDM Num Head Channels & 32 \\
VQ dropout & 0.0 & LDM Use Spatial Transformer & True \\
Condition Model & CVTP ResNet-18 & LDM Transformer Depth & 1 \\
Condition Layer & 5 & LDM Context Dim & 512 \\
Condition Frame & 5 & Batch Size & 48 \\
Cond Stage Trainable & True & Monitor & val/loss\_simple\_ema \\
Diffusion Timesteps & 1000 & Epoch & 30 \\
Scheduler & DDPM &  &  \\ \bottomrule
\end{tabular}
\end{table*}
\newpage
\section{More Qualitative Results}
We provide additional results visuo-tactile cross generation, tactile-driven stylization and tactile-driven shading estimation.
\begin{figure*}[h]
    \centering
    \raggedright
    \centering
    % \fbox{\rule{0pt}{2in} \rule{0.99\linewidth}{0pt}}
    \includegraphics[width=1.0\textwidth]{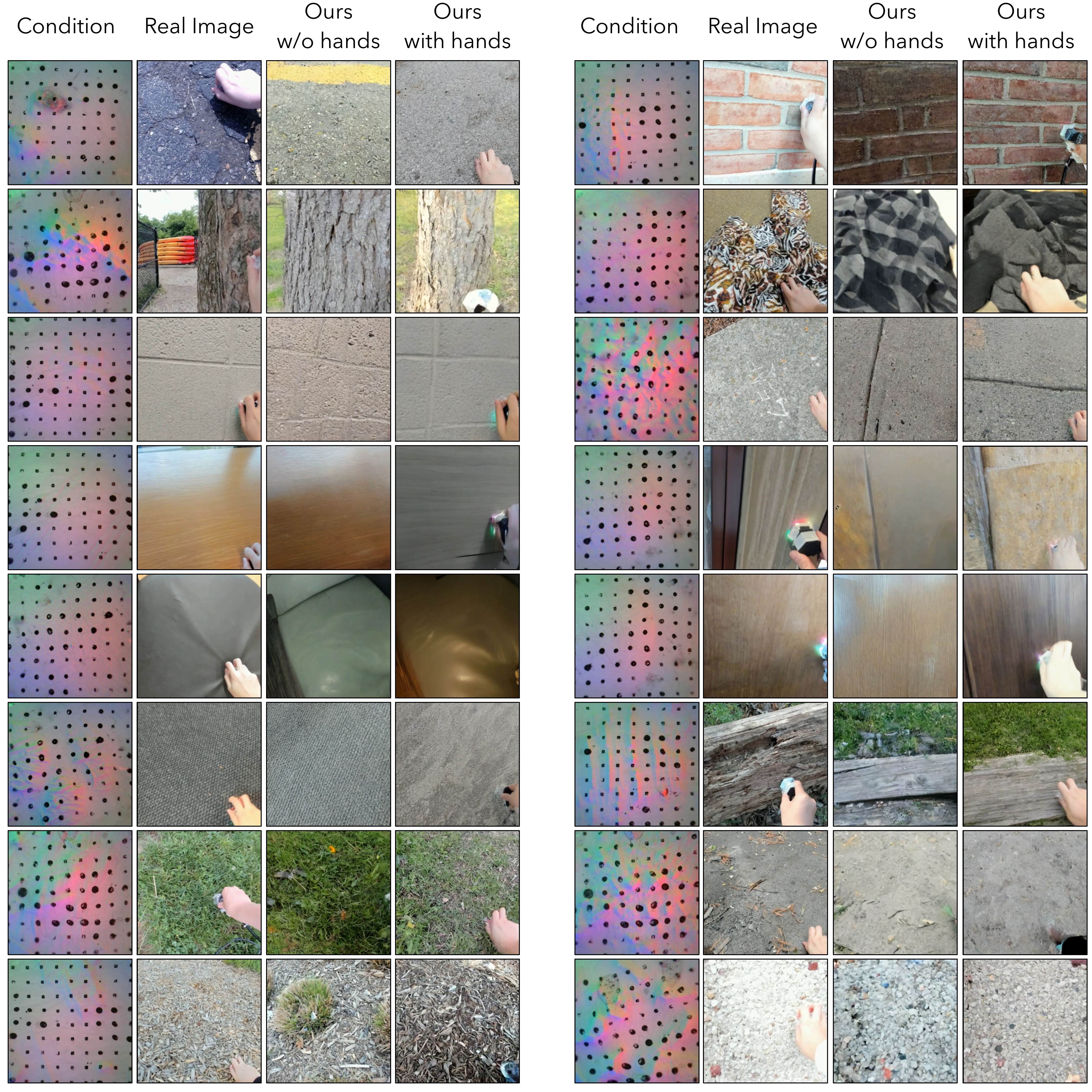}
    % \vspace{-3mm}
    \caption{Additional results for {\bf touch-to-image generation} on \emph{Touch and Go} dataset, where we show both our results with and without sensors.}
    \label{fig:pipeline}
\end{figure*}

\begin{figure*}[t]
    \centering
    \raggedright
    \centering
    % \fbox{\rule{0pt}{2in} \rule{0.99\linewidth}{0pt}}
    \includegraphics[width=1.0\textwidth]{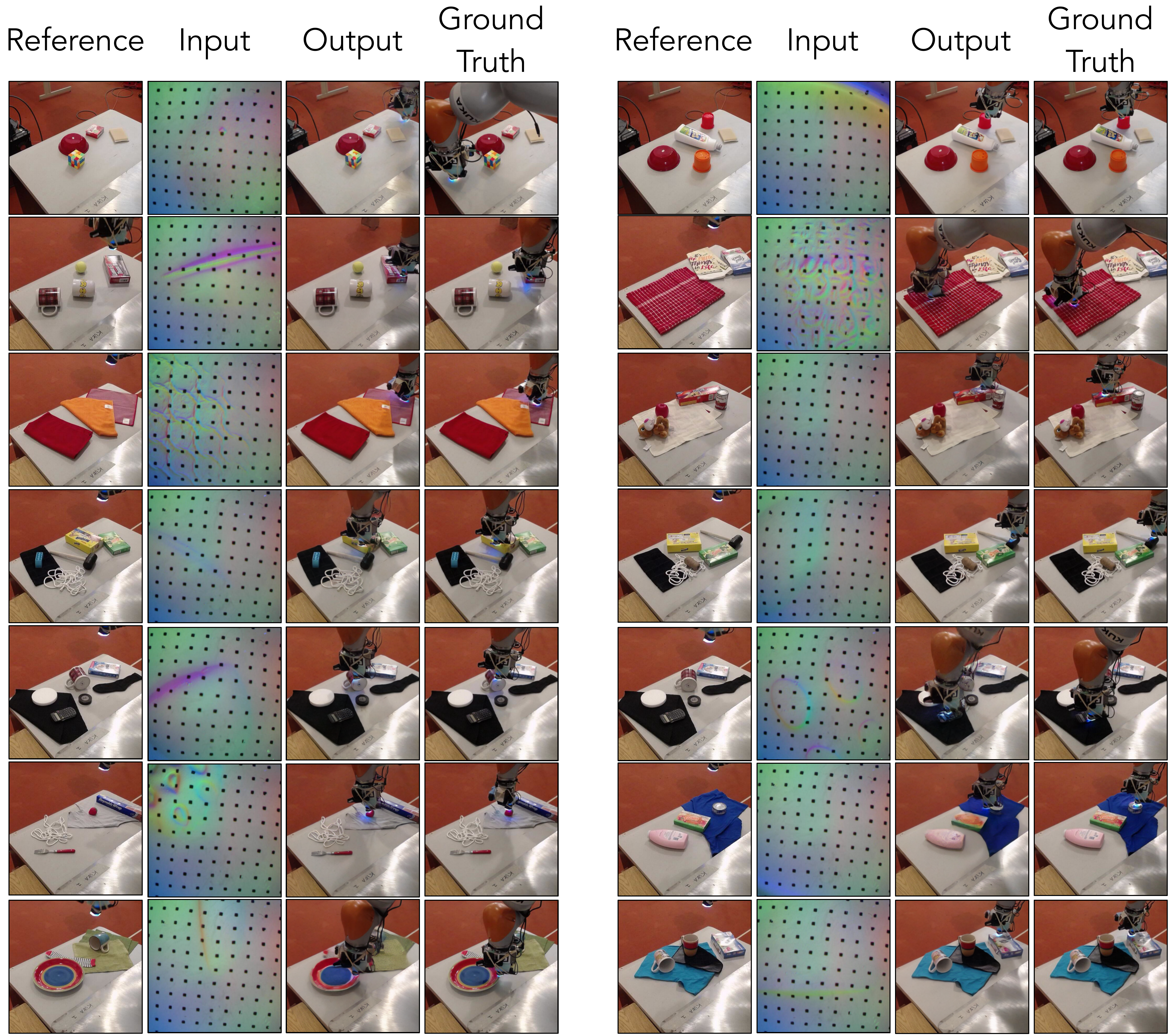}
    % \vspace{-3mm}
    \caption{Additional results for {\bf touch-to-image generation} on \emph{VisGel} dataset.}
    \label{fig:pipeline}
\end{figure*}

\begin{figure*}[t]
    \centering
    \raggedright
    \centering
    % \fbox{\rule{0pt}{2in} \rule{0.99\linewidth}{0pt}}
    \includegraphics[width=1.0\textwidth]{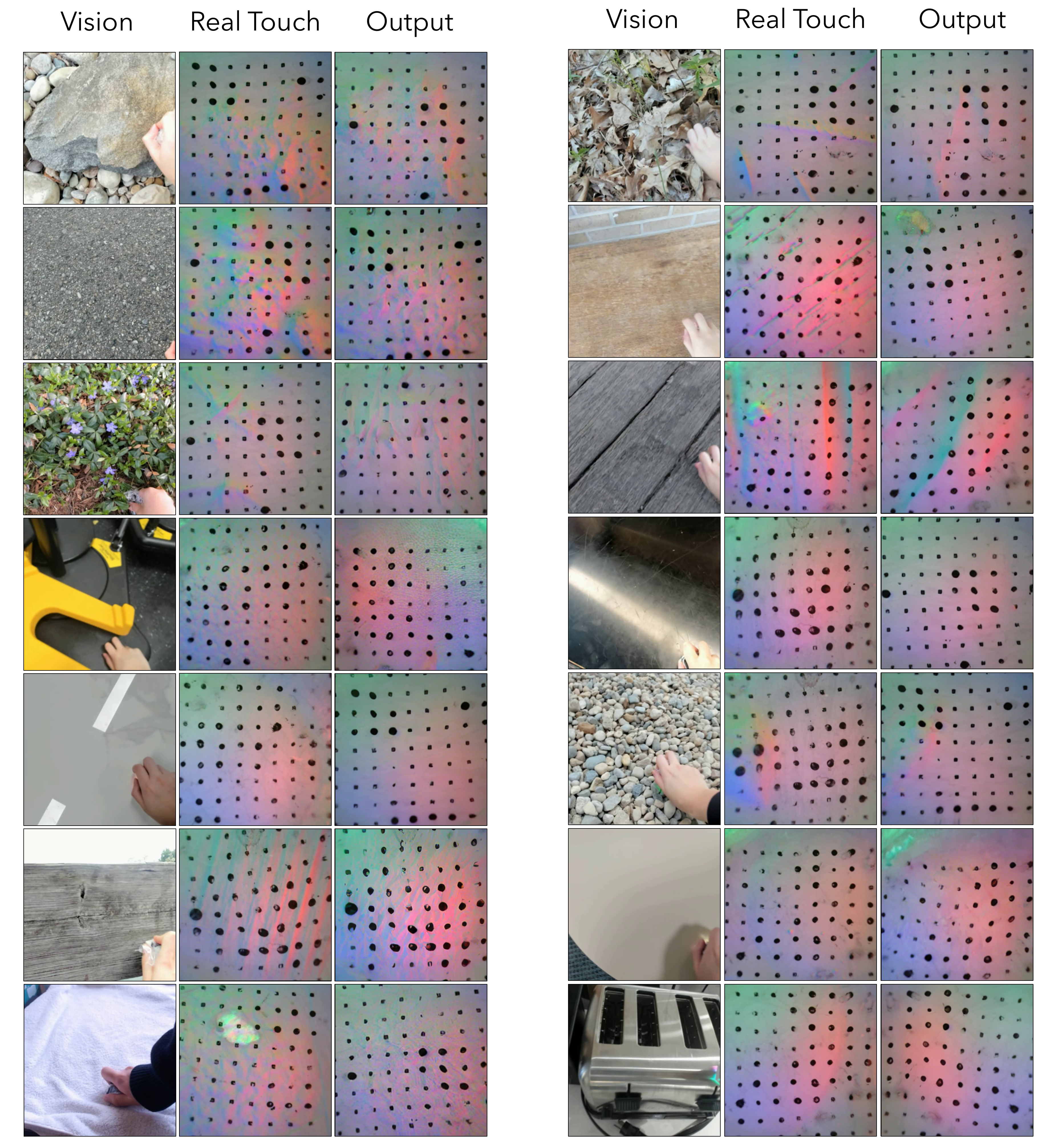}
    % \vspace{-3mm}
    \caption{Additional results for {\bf image-to-touch generation} on \emph{Touch and Go} dataset.}
    \label{fig:pipeline}
\end{figure*}

\begin{figure*}[t]
    \centering
    \raggedright
    \centering
    % \fbox{\rule{0pt}{2in} \rule{0.99\linewidth}{0pt}}
    \includegraphics[width=1.0\textwidth]{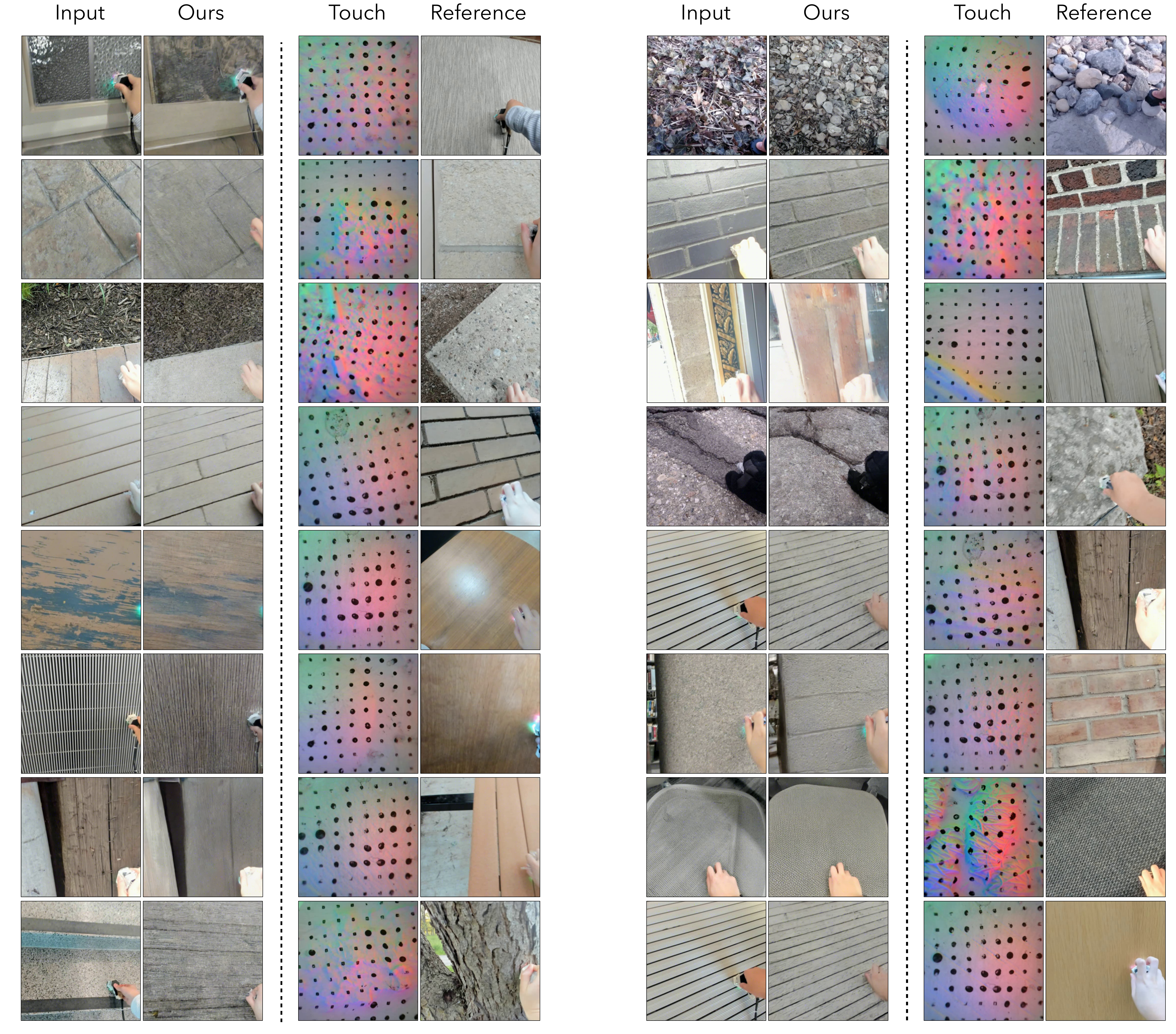}
    % \vspace{-3mm}
    \caption{Additional results for {\bf tactile-driven image stylization} results. (\emph{Zoom in for better viewing})}
    \label{fig:pipeline}
\end{figure*}

\begin{figure*}[t]
    \centering
    \raggedright
    \centering
    % \fbox{\rule{0pt}{2in} \rule{0.99\linewidth}{0pt}}
    \includegraphics[width=1.0\textwidth]{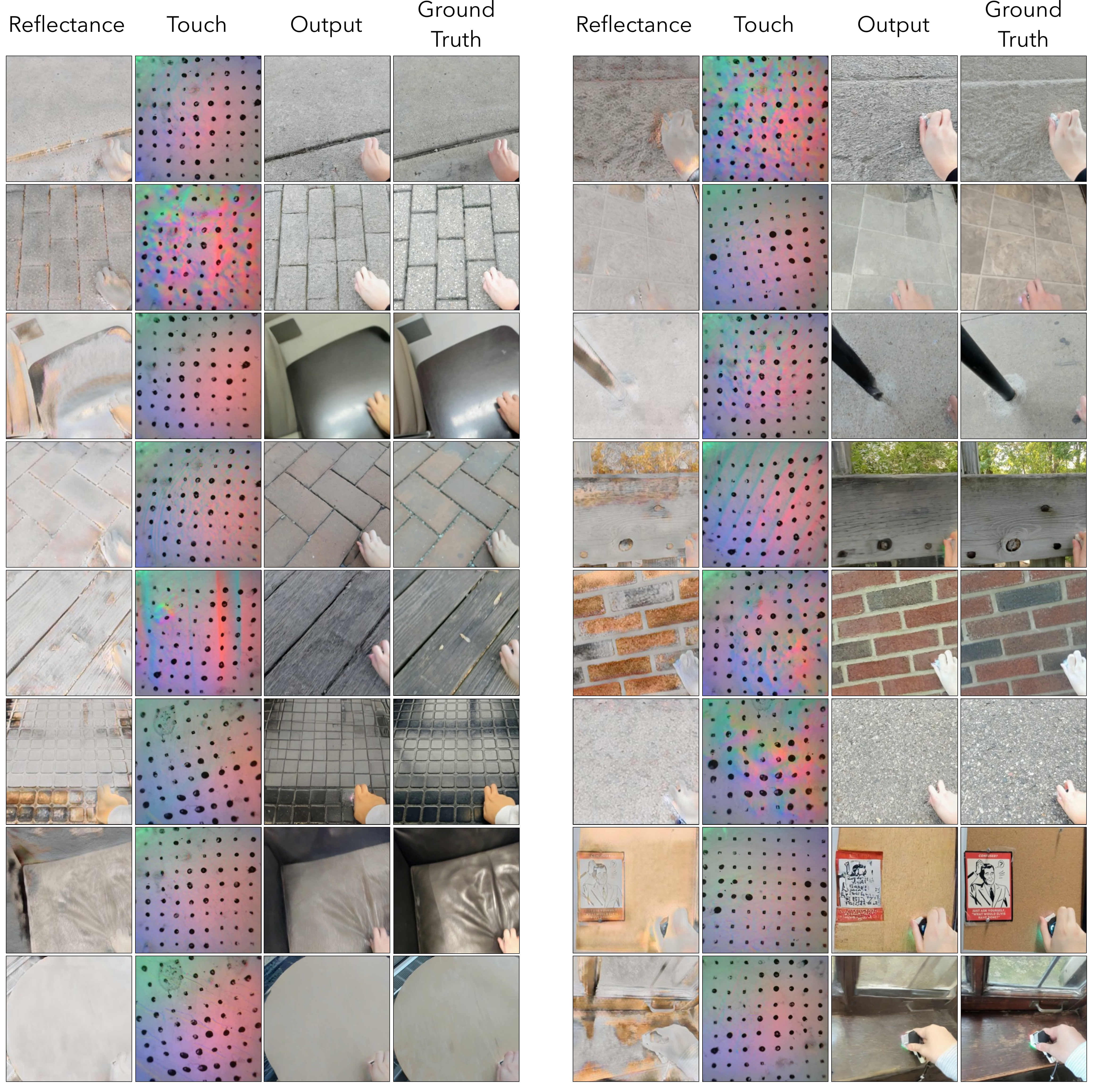}
    % \vspace{-3mm}
    \caption{Additional results for {\bf tactile-driven shading estimation}. (\emph{Zoom in for better viewing})}
    \label{fig:pipeline}
\end{figure*}

\end{document}